\documentclass[10pt,twocolumn,letterpaper]{article}

\usepackage[pagenumbers]{cvpr} 

\definecolor{cvprblue}{rgb}{0.21,0.49,0.74}
\usepackage[pagebackref,breaklinks,colorlinks,allcolors=cvprblue]{hyperref}
\usepackage{algorithm}
\usepackage{algorithmic}

\usepackage{multirow}
\usepackage{tabularray}
\usepackage{booktabs}
\usepackage{subcaption}

\def\paperID{13914} 
\def\confName{CVPR}
\def\confYear{2025}

\title{FixCLR: Negative-Class Contrastive Learning for Semi-Supervised Domain Generalization}

\author{Ha Min Son\\
University of California, Davis\\
\and
Shahbaz Rezaei\\
University of California, Davis\\
\and
Xin Liu\\
University of California, Davis\\
}
\begin{document}
\maketitle
\begin{abstract}
Semi-supervised domain generalization (SSDG) aims to solve the problem of generalizing to out-of-distribution data when only a few labels are available. 
Due to label scarcity, applying domain generalization methods often underperform. 
Consequently, existing SSDG methods combine semi-supervised learning methods with various regularization terms.
However, these methods do not explicitly regularize to learn domains invariant representations across all domains, which is a key goal for domain generalization.
To address this, we introduce FixCLR. Inspired by success in self-supervised learning, we change two crucial components to adapt contrastive learning for explicit domain invariance regularization: utilization of class information from pseudo-labels and using only a repelling term. FixCLR can also be added on top of most existing SSDG and semi-supervised methods for complementary performance improvements.
Our research includes extensive experiments that have not been previously explored in SSDG studies. These experiments include benchmarking different improvements to semi-supervised methods, evaluating the performance of pretrained versus non-pretrained models, and testing on datasets with many domains. Overall, FixCLR proves to be an effective SSDG method, especially when combined with other semi-supervised methods.
\end{abstract}    
\section{Introduction}
\label{sec:intro}

Deep learning has allowed advancement in many fields \cite{lecun2015deep}, but classification models in computer vision often struggle with predictions on data from different domains or out-of-distribution (OOD) data. Research indicates these models can fail to detect features for minority demographics \cite{benitez2022detecting, cheuk2021can} and perform poorly on domain-shifted datasets \cite{wang_imagenetSketch}. 
To address the issue of generalizing to unseen distributions, domain generalization \cite{zhou2022domain, wang2022generalizing} trains a model using data from source domains and then tests it on unseen target domains. \textcolor{black}{Note that this setup differs from \textit{domain adaptation}, where target domain data is available during training \cite{farahani2021brief}.}

A common assumption in domain generalization research is availability of fully labeled datasets. However, in real-world scenarios, fully labeled datasets are costly to collect. A more practical scenario involves having a limited number of labels and a large number of unlabeled samples. This scenario is known as semi-supervised domain generalization (SSDG), and has attracted recent attention.
Research shows that traditional domain generalization methods do not perform well in the SSDG setting, likely due to the scarcity of labels (Table \ref{tab:previous_results}). On the other hand, pseudo-labeling semi-supervised learning methods, particularly FixMatch \cite{sohn2020fixmatch}, shows promise and has become the base framework.
On top of it, StyleMatch \cite{zhou2023semistyle} adds a style-augmentation in latent space to reduce domain differences, while FBC-SA \cite{galappaththige2024towardsfbc} creates prototypes from available labeled data to maximize similarity between representations from different domains.

\begin{figure*}[t]
\centering
\begin{subfigure}{.19\textwidth}
  \centering
  \includegraphics[width=.95\linewidth,trim=2 2 2 2,clip]{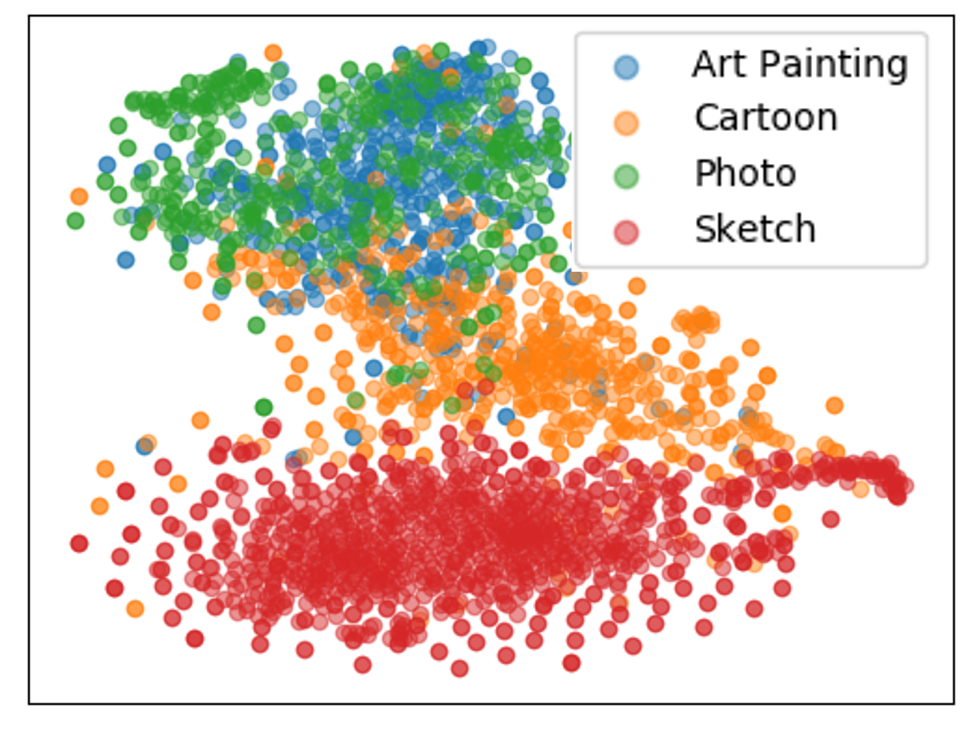}
  \caption{Original data}
  \label{fig1:sub1}
\end{subfigure}%
\begin{subfigure}{.19\textwidth}
  \centering
  \includegraphics[width=.95\linewidth,trim=2 2 2 2,clip]{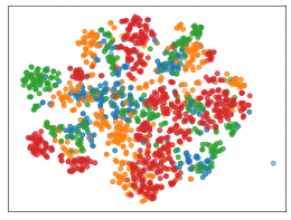}
  \caption{FixMatch}
  \label{fig1:sub3}
\end{subfigure}
\begin{subfigure}{.19\textwidth}
  \centering
  \includegraphics[width=.95\linewidth,trim=2 2 2 2,clip]{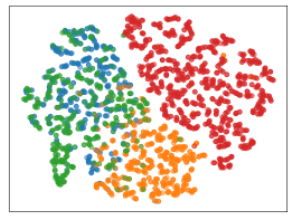}
  \caption{SimCLR}
  \label{fig1:sub2}
\end{subfigure}
\begin{subfigure}{.185\textwidth}
  \centering
  \includegraphics[width=.98\linewidth,trim=2 2 2 2,clip]{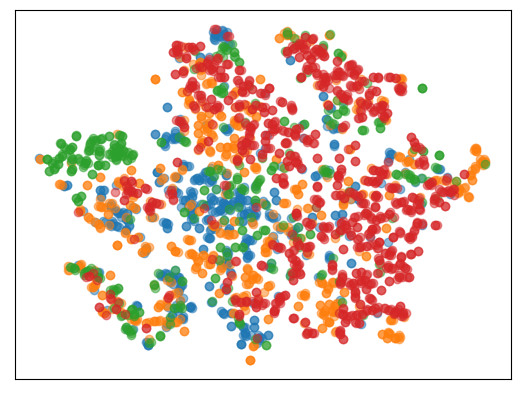}
  \caption{FixCLR(w. positives)}
  \label{fig1:positive}
\end{subfigure}
\begin{subfigure}{.188\textwidth}
  \centering
  \includegraphics[width=.98\linewidth,trim=2 2 2 2,clip]{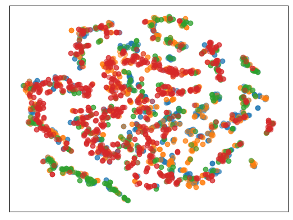}
  \caption{FixCLR}
  \label{fig1:sub4}
\end{subfigure}

\caption{t-SNE of the data manifold and the learned representation space of the learned encoder on the PACS dataset when trained with all domains. The colors represent different domains.}
\label{fig_manifold}
\end{figure*}

However, we observe one key limitation of FixMatch: FixMatch only \textit{implicitly} encourages domain invariance. The cross-entropy loss used in FixMatch  minimizes classification error, and is not incentivized to disregard domain-specific information if classification is accurate on source domains. Figure \ref{fig1:sub3} shows that FixMatch, even after fully converging, has noticeable domain clusters in the learned representation space. This is an issue for domain generalization where the goal is to learn domain invariant representations to generalize well on unseen domains \cite{zhou2022domain}.
While FBC-SA does encourage domain invariance using prototypes from each domain, only two random domains are chosen for regularization at each iteration. For datasets with many domains, FBC-SA may underperform on target domains, because regularizing two domains at a time may not be effective for achieving domain invariance across all domains.
Additionally, StyleMatch relies on an ImageNet-1K \cite{imagenet1K} pretrained network for style-augmentation, and this potentially risks domain information leakage. 

%



An unexplored approach in SSDG to explicitly encourage domain invariance with few labels is self-supervised learning \cite{chen2020simclr, grill2020byol}.
Specifically, contrastive learning methods \cite{simclrv2chen2020} have shown effectiveness in the related fields of domain adaptation, where target domain data is available \cite{singh2021clda, huang2022category}, and semi-supervised learning \cite{yao2022pcl, lewis2023improving}.
However, many self-supervised methods learn representations by attracting augmentations of the same image and repelling different images in latent space, which can lead the model to cluster domains rather than classes, as shown in Figure \ref{fig1:sub2}. Thus, unlike the related fields of domain adaptation \cite{singh2021clda, huang2022category} and fully supervised domain generalization \cite{yao2022pcl} which has access to labels, SimCLR is not directly applicable to the SSDG setup.

To the address these challenges, we introduce FixCLR, a plug-and-play regularization method that adapts contrastive learning (CLR) to explicitly regularizes for domain invariance within the FixMatch framework. Our approach is CLR-inspired, but with two crucial component changes. 
First, instead of using CLR to attract different augmentations of the same image and repel different images in latent space, FixCLR groups classes from all domains by their pseudo-labels and repels all other classes regardless of their domain. Domain invariance is encouraged by minimizing the similarity of representations between each class group and all other samples regardless of their domain. This is different from previous SSDG methods that do not explicitly regularize for domain invariance across all domains.
Second, FixCLR avoids using positive attraction for same-class samples to prevent incorrectly pseudo-labeled samples from being encoded closer together, as this could lead to the model becoming confidently incorrect and not properly converging as shown in Figure \ref{fig1:positive}. \textcolor{black}{The use of positive attraction has also been reported to cause generalization degradation in fully supervised domain generalization \cite{yao2022pcl}, which we confirm in the SSDG setting.} 

The effect of FixCLR is shown in Figure \ref{fig1:sub4}. Domain invariance is effectively learned, as shown by the absence of domain clusters. This is advantageous for generalizing to unseen domains because the model can disregard domain information and focus only on class information. 
An added benefit to the design of FixCLR is its improvement to pseudo-label quality in the SSDG setting. Traditional semi-supervised learning methods improve pseudo-label quality by correcting biases in predicted probability distributions \cite{wang2022debiasPL, schmutz2022defixmatch, chen2023softmatch}. In contrast, FixCLR regularizes for domain invariance
in the representation space thereby improving quality. 
The improvement can partially be explained with Figure \ref{fig_manifold}. When using only cross-entropy loss, as in FixMatch (Figure \ref{fig1:sub3}), some domain clustering occurs, showing that high classification accuracy can be achieved without learning domain invariant representations. However, with FixCLR (Figure \ref{fig1:sub4}), the additional regularization term changes the representation space structure and shows much less domain clustering, suggesting that the constraint changes the effect of cross-entropy. As with many regularizations, it is likely that the influence of cross-entropy is weakened, which lowers model confidence, and leads to the use of pseudo-labels from only the more confidently classifiable samples for training. Figures 2a and 2b support this observation, as pseudo-label quality improves while quantity decreases.



In summary, our contributions are as follows:
\begin{description}
  \item[$\bullet$] We introduce FixCLR, a simple yet effective novel regularization for SSDG that adapts contrastive learning, which explicitly regularizes for domain invariant representation learning across all domains (Figure \ref{fig1:sub4}).
  \item[$\bullet$] By design, FixCLR improves pseudo-label quality by encouraging domain invariance in representation space, different from traditional probability distribution debiasing methods (Figure \ref{fig2:sub1}). 
  \item[$\bullet$] We conduct extensive experiments not covered in previous SSDG research. First, we benchmark methods that improve both \textit{quality} and \textit{quantity} of pseudo-labels in FixMatch. Second, we compare performance of both pretrained and non-pretrained models. Third, we include datasets with multiple domains.
  \item[$\bullet$] FixCLR is plug-and-play and can be added on top of virtually any SSDG methods. Our extensive experiments demonstrate that FixCLR provides significant improvements in terms of accuracy and
computational efficiency.
\end{description}
\section{Related Work}
\label{sec:related}

\textbf{Domain Generalization. } 
\textcolor{black}{Domain generalization aims to train a model using data from known source domains with the objective of performing well on unseen target domains. This setting is different from domain adaptation, where target domains data are available during training time \cite{farahani2021brief}.}
Many domain generalization methods use regularization terms to encourage domain invariant representations. Methods like CrossGrad \cite{shankar2018crossgrad} and DDAIG \cite{zhou2020DDAIG} achieve this by augmentating data in the direction of greatest domain change. Similar to our work, SelfReg \cite{kim2021selfreg} uses self-supervised contrastive learning to encourage domain invariance. However, SelfReg does not use negative-class repelling, and only uses same-class different-domain attraction. In contrast, our approach relies on repelling negative-class any-domain due to the unreliability of labels.

While these methods are effective when all labels are available, they are less effective in the semi-supervised setting with limited labels \cite{zhou2023semistyle, galappaththige2024towardsfbc}. For details, we point readers to comprehensive surveys \cite{zhou2022domain, wang2022generalizing}.

\noindent\textbf{Semi-supervised learning. }
There are several types of semi-supervised learning methods relevant to our work. 
Transfer learning based methods were popularized by SimCLRv2 \cite{simclrv2chen2020} which used contrastive learning on the entire dataset before fine-tuning the classification head on few labeled data, achieving better performance than most other semi-supervised methods. However, SimCLRv2 cannot be directly applied to SSDG as it learns clusters of domains rather than clusters of classes (Figure \ref{fig1:sub2}).
FixCLR makes crucial changes to adapt it to the SSDG setting.

Entropy minimization methods \cite{grandvalet2004entropy_1, lee2013entropy_2} and consistency regularization methods \cite{sajjadi2016consistency_1, laine2016consistency_2} aim to make confident predictions on unlabeled data and ensure consistent outputs across augmented images, respectively. MixMatch \cite{berthelot2019mixmatch} combined these methods by averaging predictions from multiple augmentations to create pseudo-labels. ReMixMatch \cite{berthelot2019remixmatch} used predictions from a single weak augmentation as pseudo-labels for strongly augmented images. FixMatch \cite{sohn2020fixmatch} used a single strong augmentation alongside a weak one, generating pseudo-labels only when class confidence exceeds a threshold. FixMatch is a popular semi-supervised learning method and serves as the base SSDG framework.

To enhance pseudo-labeling in FixMatch, two main lines of work have emerged.
First, improving the quality of pseudo-labels involves addressing inaccuracies due to class imbalances in predicted probability distributions. DebiasPL \cite{wang2022debiasPL} minimizes a class-specific margin loss, which is larger for classes with lower predictions, thus promoting balanced predictions. DeFixMatch \cite{schmutz2022defixmatch} subtracts the semi-supervised loss on labeled data from the total loss to minimize the bias. 
Second, improving the quantity of pseudo-labels focuses on adjusting the threshold for considering pseudo-labels as ground truth. FreeMatch \cite{wang2022freematch} uses an exponential moving average term for global and individual class thresholds to address imbalances and uncertainty during the early stages of training. FlexMatch \cite{zhang2021flexmatch} uses Curriculum learning \cite{bengio2009curriculum} to optimize the inclusion of unlabeled data.
SoftMatch \cite{chen2023softmatch} focused on both aspects, quality and quantity, of pseudo-labels by applying uniformity to the probability distribution to encourage balanced predictions, and also includes more pseudo-labels using a softer threshold though a truncated Gaussian.

\noindent\textbf{Semi-supervised domain generalization (SSDG). }
Most relevant to our work are other SSDG research. 
Our work builds upon existing research in SSDG, particularly focusing on the scenario with constant classes and domains.

StyleMatch uses a pretrained network for style augmentation via AdaIN \cite{huang2017AdaIN} and a stochastic classifier to reduce overfitting. However, its reliance on a pretrained network for style transfer is a possible limitation. As we show in our results, a pretrained network has a significant impact on SSDG performance, potentially benefiting from  domain information leakage given the evaluation protocol. Additionally, it requires more computations, i.e. two forward passes per batch.

FBC-SA \cite{galappaththige2024towardsfbc} introduces two regularization terms: feature-based conformity (FBC) and semantic alignment (SA). FBC uses prototypes, recalculated each epoch, to measure representational similarities between unlabeled samples and prototypes from both the same and different domains. SA aligns probability distributions using cross-entropy loss to ensure consistency across domains. However, FBC-SA's reliance on meaningful prototypes and its random selection of two domains for regularization may underperform in datasets with many domains. Additionally, the method requires extra forward passes for prototype calculations, making it less efficient. In contrast, FixCLR learns to regularize for domain invariance directly in representation space for all domains, avoiding the need for style-augmentation and prototype calculations.

 \section{Proposed Method --- FixCLR}
\label{sec:method}

\textcolor{black}{Previous SSDG methods \cite{zhou2023semistyle, galappaththige2024towardsfbc} showed that traditional domain generalization methods are much less effective for the SSDG setup compared to semi-supervised methods (Table \ref{tab:previous_results}).} Following these studies, we use FixMatch \cite{sohn2020fixmatch} as our base framework. Intuitively, FixMatch and cross-entropy loss should be creating representations that cluster by class-information, and not domain information, as cross-entropy loss is minimized when the classifier effectively creates decision boundaries for each class based on learned representations. However, we find that this implicit learning of domain invariance is not very effective as seen in Figure \ref{fig1:sub3}, where clusters of domains are clearly visible. The cross-entropy loss is not incentivized to learn domain invariant representations when effective classification is possible without it. 
However, domain clusters are undesirable for domain generalization because they indicate that the model encodes domain-specific information on source domains, which is harmful when generalizing to unseen domains.


To address this issue, FixCLR \textit{explicitly} regularizes the model to learn domain invariant representations given limited labels using an adapted contrastive learning loss, motivated by the robustness of self-supervised contrastive learning. 
However, there are issues with applying SimCLR \cite{simclrv2chen2020} directly. In SimCLR, augmentations of the same image are attracted to each other, while different images are repelled. This approach encourages the model to learn representations similar to the original data manifold, leading to clusters of domains rather than classes, as shown in Figure \ref{fig1:sub2}.

FixCLR introduces two novel componential modifications to SimCLR. First, it uses class information from the predicted pseudo-labels to group together same-class samples. Second, instead of positive attraction between groups, only negative repelling is used. Same-class samples across all domains are repelled from different-class samples, regardless of domain. 
Same-class attraction is not used because cross-entropy loss already encourages the formation of class clusters to a degree. 
Therefore, additional regularization to encourage class clusters is unnecessary, particularly when pseudo-labels are uncertain. In fact, including same-class attraction can reduce the quality of pseudo-labels, as shown in Figure \ref{fig2:quality_quantity}, and ultimately reduce performance as shown in Table \ref{tab4:fixclr_positive}. These results are discussed further in a later section.


We also intentionally do not make modifications to the semi-supervised FixMatch framework, as previous SSDG research \cite{zhou2023semistyle, galappaththige2024towardsfbc} has shown its efficacy. Instead, FixCLR adds a simple yet effective regularizer to encourage domain invariance, complementing FixMatch and its improvements for pseudo-label quality and quantity. 

Formally, we define the regularization loss as follows: 

\begin{equation}
    \mathcal{L}_C = \sum_{i=0}^d -\log\frac{exp(1/\tau)}{\sum_{j=0}^c exp(sim(DOM^i_{-j}, CLS_j) / \tau)}.
    \label{eq:fixclr}
\end{equation}

\noindent Here, $\tau$ is a temperature term that is fixed to 0.5 as in SimCLR \cite{simclrv2chen2020}. The numerator is fixed to a constant because positive attraction is not used. $sim$ is the cosine similarity, $d$ is the number of domains, and $c$ is the number of classes. $DOM^i_{-j}$ is the latent space representations of any sample from the $i$\textsuperscript{th} domain that does not belong to the $j$\textsuperscript{th} class. $CLS_j$ is the latent space representation of all samples in the $j$\textsuperscript{th} class, regardless of domain.
The final loss function for FixCLR is defined as $\mathcal{L} = \mathcal{L}_S + \mathcal{L}_U + \mathcal{L}_C$. Here, $\mathcal{L}_S$ is the cross-entropy loss of the labeled data, and $\mathcal{L}_U$ is the cross-entropy loss of the unlabeled data as defined in FixMatch \cite{sohn2020fixmatch}.

FixCLR explicitly regularize the model to learn domain invariant representations by grouping samples predicted to belong to the same class, then minimizing the representational similarity between the same-class groups and different-class any-domain groups. Consequently, the regularization loss is minimized when the model does not encode domain-specific representations.
The effect of FixCLR is shown Figure \ref{fig1:sub4}, where clusters of domains are absent, unlike the original data manifold and other methods. Instead, samples are intermixed regardless of domain, showing that domain information is disregarded in the encoding process. On unseen domains, the model trained with FixCLR is more likely to focus less on domain information, focusing more on features that differentiate classes, aligning with the goal of domain generalization.

\begin{table}
\centering
\fontsize{10pt}{10pt}\selectfont
\setlength{\tabcolsep}{5pt} 
\renewcommand{\arraystretch}{1.0} 
\color{black}
\begin{tabular}{cccccc}
\toprule

& Model & Digits  & PACS  & OH  & Terra \\
 \midrule
\multirow{3}{*}{DG} & ERM         & 29.1\% & 59.8\% & 56.7\% & 23.5\% \\
                    & EntMin      & 39.3\% & 57.0\% & 57.0\% & 26.6\% \\
                    & MeanTeacher & 38.8\% & 55.9\% & 55.9\% & 25.0\% \\
\midrule
\multirow{3}{*}{SSL} & FlexMatch  & 68.9\% & 72.7\% & 53.7\% & 26.4\% \\
                     & FreeMatch  & 67.5\% & 74.0\% & 56.2\% & 30.1\% \\
                     & FixMatch   & 66.4\% & 76.6\% & 57.8\% & 30.5\% \\
\bottomrule
\end{tabular}
\caption{\textcolor{black}{Results from \cite{galappaththige2024towardsfbc} (CVPR 2024) showing traditional domain generalization methods (DG) underperform compared to semi-supervised methods (SSL) in the semi-supervised domain generalization setup.}}
\label{tab:previous_results}
\end{table}

\begin{table*}
\renewcommand{\arraystretch}{1.15}

\centering
\fontsize{9pt}{9pt}\selectfont
\begin{tabular}{cccccccc} 
\toprule
~                           & Method / Dataset     & Digits DG & PACS & Office-Home   & Terra Inc. & ImageNet-R & FMOW-Wilds  \\ 
\hline
~ & FixMatch   & 60.3   & 76.0   & 56.3 & 28.3 & 24.0 & 24.5  \\ 
\hline
\multirow{2}{*}{Quality}    & DebiasPL   & 60.3   & 75.8 & 56.3 & 25.0 & 23.8 & 19.9  \\
                            & DeFixMatch & 35.3   & 77.0 & 57.5 & 28.0 & \underline{24.8} & 19.3  \\ 
\hline
\multirow{2}{*}{Quantity}   & FlexMatch  & 60.5   & 77.8 & 57.0   & 25.8 & 23.3 & 22.1  \\
                            & FreeMatch  & 61.8   & 77.8 & 57.5   & 24.0 & 23.8 & 21.0  \\ 
\hline
Quality + Quantity          & SoftMatch  & 64.8   & 75.5 & 57.8   & 22.0 & 24.1 & 18.4  \\ 
\hline
\multirow{3}{*}{SSDG}       & StyleMatch & 58.0   & \underline{78.3} & \underline{59.0}   & 24.3 & 24.5 & \underline{24.9}  \\
                            & FBC-SA     & 67.5   & 77.0 & \underline{59.0}   & 30.8 & 23.3 & -     \\
                            & FixCLR     & \underline{67.8}   & 76.0 & 57.8   & \underline{33.8} & 24.0 & 24.5  \\ 
\hline \hline
\multicolumn{8}{c}{Improvements when FBC-SA/FixCLR added}                     \\ 
\hline
\multirow{2}{*}{DebiasPL}   & +FBC-SA    & \textbf{+7.5} & \textbf{+2.0} & \textbf{+3.0}  & +3.0 & -8.0 & -     \\
                            & +FixCLR    & +7.2 & +0.7 & +1.2  & \textbf{+7.0} & \textbf{+1.5} & -2.4  \\ 
\hline
\multirow{2}{*}{DeFixMatch} & +FBC-SA    & -13.3& -0.2 & \textbf{+2.3}  & -6.7 & \textbf{-0.9} & -     \\
                            & +FixCLR    & \textbf{-10.3}& \textbf{+1.5}  & +1.3  & \textbf{+2.5} & -1.6 & +4.0   \\ 
\hline
\multirow{2}{*}{FlexMatch}  & +FBC-SA    & \textbf{+7.0} & \textbf{-1.3} & \textbf{+2.3}  & +0.7 & -5.9 & -7.0    \\
                            & +FixCLR    & +6.3 & -1.5 & +0.8  & \textbf{+8.0} & \textbf{+1.6} & \textbf{+0.7}   \\ 
\hline
\multirow{2}{*}{FreeMatch}  & +FBC-SA    & \textbf{+7.2} & -1.3 & \textbf{+1.8}  & +3.3  & 0.0 & -0.4  \\
                            & +FixCLR    & +6.5 & \textbf{-0.5} & +0.3  & \textbf{+10.0} & \textbf{+0.5} & \textbf{+0.0} \\ 
\hline
\multirow{2}{*}{SoftMatch}  & +FBC-SA    & +7.7 & \textbf{+1.5} & +2.0  & +2.3  & -0.7 & -     \\
                            & +FixCLR    & \textbf{+8.2} & +1.3 & \textbf{+2.2}  & \textbf{+13.5} & \textbf{+1.8} & +3.9   \\ 
\midrule
\multirow{2}{*}{StyleMatch} & +FBC-SA    & \textbf{+6.3} & +0.5 & +0.3  & +0.7  & +0.1 & -     \\
                            & +FixCLR    & +5.8 & \textbf{+0.7} & \textbf{+3.3}  & \textbf{+9.2}  & \textbf{+0.2} & +0.5   \\
\bottomrule
\end{tabular}
\caption{Performance on various datasets with an \textbf{ImageNet pretrained} network. We \underline{underline} the best \underline{\textbf{standalone}} method, and \textbf{bold} the best improvement for each method when FBC-SA / FixCLR are added.}
\label{tab1:pretrained}
\end{table*}

\begin{table}
\centering
\fontsize{9pt}{9pt}\selectfont
\setlength{\tabcolsep}{1.15pt} 
\renewcommand{\arraystretch}{1.0} 
\begin{tabular}{l|cccccc}
\toprule
Dataset & Digits  & PACS  & OH  & Terra  & IMG-R & FMOW \\
 \midrule
\begin{tabular}[l]{@{}l@{}}Best \\Method\end{tabular}  & \begin{tabular}[c]{@{}l@{}}FixCLR \\+ Soft\end{tabular} & \begin{tabular}[c]{@{}l@{}}FixCLR \\+ Style\end{tabular} & \begin{tabular}[c]{@{}l@{}}FixCLR \\+ Style\end{tabular} & \begin{tabular}[c]{@{}l@{}}FixCLR \\+ Soft\end{tabular} & 
\begin{tabular}[c]{@{}l@{}}FixCLR \\+ Soft\end{tabular} & \begin{tabular}[c]{@{}l@{}}FixCLR \\+ Style\end{tabular}  \\

\\

Accuracy & 73.0\%  & 79.0\%  & 62.3\%  & 35.5\%  & 25.9\% & 25.4\%
\\ \bottomrule
\end{tabular}
\caption{Highest performing method on the ImageNet pretrained network}
\label{tab:highest_pre}
\end{table}

\subsection{Experimental Setup}
\noindent \textbf{Datasets and architecture. }
We use six widely used domain generalization datasets. We include three relatively simple datasets. Digits DG \cite{zhou2020digitsDG} has 4 domains and 10 classes, PACS \cite{li2017pacs} 4 domains and 7 classes, Office-Home \cite{venkateswara2017officehome} has 4 domains and 62 classes. A slightly more complex dataset is Terra Incongnita \cite{beery2018terra} due to its image style with 4 domains and 10 classes. 
Importantly, we also include two complex datasets with more domains not used in prior SSDG research. 
Imagenet-R(endition) \cite{hendrycks2021ImagenetR} has 15 domains and 200 classes. 
FMOW-Wilds \cite{koh2021fmowwilds} has 13 domains and 62 classes. The FMOW-Wilds is the most complex, as it is a hybrid subpopulation and DG dataset \cite{koh2021fmowwilds}.
For the Digits DG, PACS, Office-Home, and Terra Incognita, we follow the leave-one-domain-out evaluation, where one domain is used as the target domain while the remaining three domains are used as source domains for training. The target domain remains unseen until test time. We report the average accuracy of the target domain across all four runs. 
For ImageNet-R, we select the `misc' domain as the target domain as it contains images from many other unseen domains. For the FMOW-Wilds dataset, we use the predefined test set \cite{koh2021fmowwilds} as the target domain. We report the average accuracy across four runs. 
For all datasets, we run experiments with both an ImageNet pretrained network and non-pretrained network. Furthermore, following \citet{zhou2023semistyle, galappaththige2024towardsfbc} we run experiments with 10 labels and 5 labels per class. We report the 10 label per class setting except for ImageNet-R which as 200 classes. We report the 5 label per class for ImageNet-R. The 5-label case for other datasets are included in the supplementary materials due to space limitations.

Following \citet{zhou2023semistyle, galappaththige2024towardsfbc} we use a ResNet-18 \cite{targ2016resnet} for all experiments, and set the batch size to 48 for the datasets with four domains. For ImageNet-R we use a batch size of 42, and for we use FMOW-Wilds we use a batch size of 44. We use different batch sizes to accommodate FBC-SA which requires the batch size to be divisible by the number of source domains.
We also use SGD as the optimizer with an initial learning rate of 0.003 and reduce this with the cosine annealing rule. We train for 20 epochs on all datasets except Terra Incognita which is trained for 10 epochs, following \citet{galappaththige2024towardsfbc}, and ImageNet-R which is trained for 50 epochs.

\noindent \textbf{Baselines. }
As previous SSDG methods \cite{zhou2023semistyle, galappaththige2024towardsfbc} showed that semi-supervised learning methods are the most effective framework for SSDG (Table \ref{tab:previous_results}), we use it as our main framework and add additional loss terms to improve performance.
We also evaluate methods that improve pseudo-labeling in FixMatch. For quality improvements, we include DeFixMatch \cite{schmutz2022defixmatch} and DebiasPL \cite{wang2022debiasPL}. For quantity improvements, we include FreeMatch \cite{wang2022freematch} and FlexMatch \cite{zhang2021flexmatch}. SoftMatch \cite{chen2023softmatch} addresses both quality and quantity.

Additionally, we compare against SSDG methods, namely StyleMatch \cite{zhou2023semistyle} and FBC-SA \cite{galappaththige2024towardsfbc}. Similar to \citet{galappaththige2024towardsfbc}, we 
add our domain invariance regularization terms to the different semi-supervised methods. Unlike FBC-SA, we also include methods that target pseudo-label quality.
All experiments were conducted on a single RTX A5000 and one AMD EPYC 7763 processor.

\section{Results and Discussion}
\label{sec:results}

\begin{table*}
\renewcommand{\arraystretch}{1.15}

\centering
\fontsize{9pt}{9pt}\selectfont
\begin{tabular}{cccccccc} 
\toprule
~                           & Method / Dataset     & Digits DG & PACS & Office-Home   & Terra Inc. & ImageNet-R & FMOW-Wilds  \\ 
\hline
~ & FixMatch           & 56.5   & 37.0   & 19.8        & 16.5         & 6.9        & 14.9       \\
\hline
\multirow{2}{*}{Quality}    & DebiasPL & 57.0  & 34.3 & 20.5 & 16.5 & 7.4        & 16.0         \\
                            & DeFixMatch         & 26.3      & 35.5 & 19.5        & 16.8         & 6.7        & 13.3       \\
\hline
\multirow{2}{*}{Quantity}   & FlexMatch          & 58.0      & 36.8 & 20.3        & 17.8         & 7.1        & 14.4       \\
                            & FreeMatch          & 60.5      & 37.8 & 20.3        & 17.5         & 7.0        & 15.0         \\
\hline
Quality + Quantity          & SoftMatch          & \underline{62.3} & 37.0 & \underline{20.8}        & 18.0  & 6.8        & 16.3       \\
\hline
\multirow{3}{*}{SSDG}       & StyleMatch         & 56.3      & \underline{37.5} & \underline{20.8}        & 15.8         & \underline{7.9} & \underline{18.1}  \\
                            & FBC-SA             & 59.5      & 37.8 & 20.0        & 16.5         & 7.4        & 14.1       \\
                            & FixCLR             & 58.0      & 35.3 & 20.3        & \underline{19.5}  & 7.2        & 15.6       \\
\hline \hline
\multicolumn{8}{c}{Improvements when FBC-SA/FixCLR added}                     \\ 
\hline
\multirow{2}{*}{DebiasPL}   & +FBC-SA            & \textbf{+5.0}      & \textbf{+3.7} & -0.5        & \textbf{+1.0}           & \textbf{+0.1}       & -          \\
                            & +FixCLR            & +1.0      & +1.5 & \textbf{-0.2}        & +0.0          & -0.4       & -1.4       \\
\hline
\multirow{2}{*}{DeFixMatch} & +FBC-SA            & \textbf{+1.2}      & \textbf{+1.3} & +0.0        & +4.0           & +0.5       & -          \\
                            & +FixCLR            & -3.5      & -1.5 & \textbf{+1.0}        & \textbf{+4.1}         & \textbf{+0.6}       & +1.5       \\
\hline
\multirow{2}{*}{FlexMatch}  & +FBC-SA            & \textbf{+3.0}      & +1.2 & -0.3        & +0.2         & -0.1       & +0.8       \\
                            & +FixCLR            & +1.0      & +1.2 & \textbf{+0.2}         & \textbf{+1.5}         & \textbf{+0.3}       & \textbf{+1.3 }      \\
\hline
\multirow{2}{*}{FreeMatch} & +FBC-SA            & \textbf{-0.2}      & \textbf{+0.2} & +0.0         & +1.8         & +0.5       & \textbf{+0.2}       \\
                           & +FixCLR            & -0.7      & -2.8 & +0.0         & \textbf{+1.8}         & \textbf{+0.6}       & -0.1       \\
\hline
\multirow{2}{*}{SoftMatch}  & +FBC-SA            & \textbf{+3.5}      & +1.8 & \textbf{+1.2}         & +1.3         & +0.2       & -          \\
                            & +FixCLR            & +1.6      & \textbf{+5.5} & +1.0         & \textbf{+2.3}         & \textbf{+0.5}       & -0.2       \\
\midrule
\multirow{2}{*}{StyleMatch} & +FBC-SA            & \textbf{+5.0}      & +0.8 & -0.3        & -0.3         & +0.0       & -          \\
                            & +FixCLR            & +1.5      & \textbf{+3.7} & \textbf{+1.5}         & \textbf{+2.7}         & \textbf{+0.4}       & +0.5       \\
\bottomrule
\end{tabular}
\caption{Performance on various datasets with a \textbf{non-pretrained} network. We \underline{underline} the best \underline{\textbf{standalone}} method, and \textbf{bold} the best improvement for each method when FBC-SA / FixCLR are added.}
\label{tab2:nonpretrained}
\end{table*}

\begin{table}
\centering
\fontsize{9pt}{9pt}\selectfont
\setlength{\tabcolsep}{1.15pt} 
\renewcommand{\arraystretch}{1.0} 
\begin{tabular}{l|cccccc}
\toprule
Dataset & Digits  & PACS  & OH  & Terra  & IMG-R & FMOW \\
 \midrule
\begin{tabular}[l]{@{}l@{}}Best \\Method\end{tabular}  & 
\begin{tabular}[c]{@{}l@{}}FBC-SA \\+ Soft\end{tabular} & 
\begin{tabular}[c]{@{}l@{}}FixCLR \\+ Soft\end{tabular} & 
\begin{tabular}[c]{@{}l@{}}FixCLR \\+ Style\end{tabular} & 
\begin{tabular}[c]{@{}l@{}}FixCLR \\+ DeFix\end{tabular} & 
\begin{tabular}[c]{@{}l@{}}FixCLR \\+ Style\end{tabular} & 
\begin{tabular}[c]{@{}l@{}}FixCLR \\+ Style\end{tabular}  \\

\\

Accuracy & 65.8\%  & 42.5\%  & 22.3\% & 20.9\% & 8.3\% &  18.6\%
\\ \bottomrule
\end{tabular}
\caption{Highest performing method on the non-pretrained network}
\label{tab:highest_non}
\end{table}

\textbf{Combining SSDG and semi-supervised methods. }
A key objective of this paper is to demonstrate that FixCLR is a simple yet performant method for SSDG in terms of accuracy. Table \ref{tab1:pretrained} shows extensive results on various datasets as using an ImageNet pretrained network, consistent with previous SSDG research \cite{zhou2023semistyle, galappaththige2024towardsfbc}. 

We find that relying on SSDG methods (top half of table) does not always yield the best performance. Improvements to the quality and quantity of pseudo-labels fairly consistently improve performance, which aligns with our expectations. Prior research has shown that semi-supervised methods, especially FixMatch, outperform domain generalization methods in the SSDG setting. Therefore, we included both improvements to pseudo-label quality and quantity, expecting them to be strong baselines. 

FixCLR focuses on regularizing domain invariant representations across all domains rather than improving pseudo-labeling. Consequently, we ran experiments combining pseudo-label improvement methods with FixCLR (bottom half of table), following a similar approach to \citet{galappaththige2024towardsfbc}. We also tested combinations with FBC-SA for a fair comparison. We did not include StyleMatch in these combinations because its use of a pretrained network for style-augmentation may provide an unfair advantage. 

We find that combining FixCLR and FBC-SA with semi-supervised methods often leads to complementary improvements performance in the SSDG setup. The improvement in performance is especially higher for the Digits DG and Terra Incognita datasets, where FixCLR and FBC-SA perform well individually. The significant improvements for specific datasets indicate that encouraging domain invariance can be more important in certain datasets. 

\noindent\textbf{FixCLR versus FBC-SA. }
When combined with different methods, FixCLR consistently leads to larger improvements compared to FBC-SA. This suggests that FixCLR is more effective at encouraging domain invariance when combined with other methods. Although the improvements do not significantly differ in performance for the simpler Digits DG, PACS, and Office-Home datasets, the differences are signficiant with more complex datasets. 
Table \ref{tab:highest_pre} shows the highest overall performing methods, including the combinations. FixCLR, in combination with SoftMatch or StyleMatch, consistently achieves the best results. 

Additionally, FBC-SA struggles with many-domain datasets, such as ImageNet-R and FMOW-Wilds. For ImageNet-R, FBC-SA consistently leads to worse performance, likely because its reliance on choosing two random domains to maximize similarity does not effectively encourage domain invariance when there are many domains. 
In the FMOW-Wilds dataset, which is also a hybrid subpopulation task, FBC-SA collapses during training. 

\noindent\textbf{Pretrained versus non-pretrained models. }
Examining Table \ref{tab2:nonpretrained}, which presents results using non-pretrained models, reveals significant differences in performance compared to Table \ref{tab1:pretrained}. 
The performance gap is especially large for datasets that have visually similar images to ImageNet (PACS, Office-Home, and ImageNet-R). As it is standard to perform leave-one-domain-out evaluation, pretrained models may have an advantage in such cases as they have been exposed to similar domain information during pretraining. 

Of course, there are clear benefits to using pretrained models and this has already been reported in previous semi-supervised research \cite{xu2024revisiting}, domain generalization research \cite{yu2024rethinking}, and also observed in our results for datasets like Digits DG, Terra Incognita, and FMOW-Wilds which are less visually similar to ImageNet. However, there is a need to quantify domain information leakage. This would allow fair comparisons across methods. For instance, StyleMatch uses an ImageNet pretrained VGG network for style-augmentation, which may provide an unfair advantage when training a model from scratch, especially if the target domain has similarities to ImageNet. 

Despite these concerns, when using a non-pretrained network, FixCLR combined with other methods perform best for most datasets as shown in Table \ref{tab:highest_non} similar to the pretrained network case. 
The 5-label case show similar findings (supplementary materials).

\noindent\textbf{Pseudo-Label Quality and Quantity for SSDG. }
Previous SSDG research reported on the performance benefits of FixMatch in the SSDG setup \cite{galappaththige2024towardsfbc, zhou2023semistyle}. However, comparisons focusing on improvements to FixMatch pseudo-labels, specifically in terms of quality and quantity, were not conducted. We expected that SoftMatch \cite{chen2023softmatch} which addresses both quality and quantity aspects, would outperform other semi-supervised methods.

Our findings shown in Table \ref{tab:highest_pre} and Table \ref{tab:highest_non} support our expectation to an extent. SoftMatch combined with FixCLR is often the highest performing method.
However, when SoftMatch is used alone, results were inconsistent. With pretrained models (Table \ref{tab1:pretrained}), SoftMatch only outperformed other methods on Digits DG and Office-Home. In contrast, with non-pretrained models (Table \ref{tab2:nonpretrained}), SoftMatch performed better on four out of six datasets. 


We also observed interesting results with FMOW-Wilds. Neither quality nor quantity approaches outperformed StyleMatch, with most methods performing worse than the base FixMatch when using a pretrained network (Table \ref{tab1:pretrained}). This underperformance is likely due to class imbalances in the target domain. FMOW-Wilds is a hybrid task focused also on high performance across minority classes. StyleMatch likely performed best because it is the only method that uses a stochastic classifier, which reduces overfitting to majority classes.

Overall, it is difficult to conclude which pseudo-label improvements in quality or quantity are best for SSDG due to the inherant quality-quantity trade-off. Some datasets benefit more when only quality improvements (DebiasPL/DeFixMatch) are used,
while others benefit more when only quantity improvements (FlexMatch/Freematch) are used. When combined with FixCLR, SoftMatch--addressing both quality and quantity--typically performs best. Therefore, on different datasets it is advisable to test various methods to find which approach achieves best performance, especially if FixCLR is not used.

\noindent\textbf{Efficiency. }
Another advantage of FixCLR is computationally efficiency. Unlike previous SSDG methods, FixCLR does not require additional forward passes. StyleMatch requires two forward passes per batch to create a style-augmented image, while FBC-SA requires a forward pass over all labeled data to find prototypes per epoch.
As a result, FixCLR is more efficient in terms of wall clock time during training, as shown in Table \ref{tab3:efficiency}. Even when combined with other methods, FixCLR remains more efficient compared to FBC-SA. Note that FixMatch is the baseline as all methods add some form of regularization to FixMatch.

\begin{table}
\centering
\fontsize{9pt}{9pt}\selectfont
\setlength{\tabcolsep}{2.0pt} 
\renewcommand{\arraystretch}{1.0} 
\begin{tabular}{l|cccccc}
\toprule
Method/Dataset & Digits  & PACS  & OH  & Terra  & IMG-R & FMOW \\
 \midrule
FixMatch    & 20  & 26  & 60  & 201 & 248 & 582 \\
StyleMatch  & 60  & 123 & 167 & 528 & 651 & 1162\\
FBC-SA      & 31  & 42  & 97  & 267 & 294 & 671\\
FixCLR      & 26  & 36  & 75  & 221 & 263 & 610 \\ 
Soft+FBC-SA & 35  & 50  & 112 & 279 & 302 & 693\\
Soft+FixCLR & 30  & 41  & 83  & 229 & 283 & 623\\
Style+FBC-SA& 67  & 56  & 181 & 608 & 697 & 1283\\
Style+FixCLR& 63  & 56  & 173 & 559 & 672 & 1206\\
\bottomrule
\end{tabular}
\caption{Average time (in seconds) per epoch}
\label{tab3:efficiency}
\end{table}

\begin{figure}[t]
\centering
\begin{subfigure}{.22\textwidth}
  \centering
  \includegraphics[width=.98\linewidth,trim=3 3 3 3,clip]{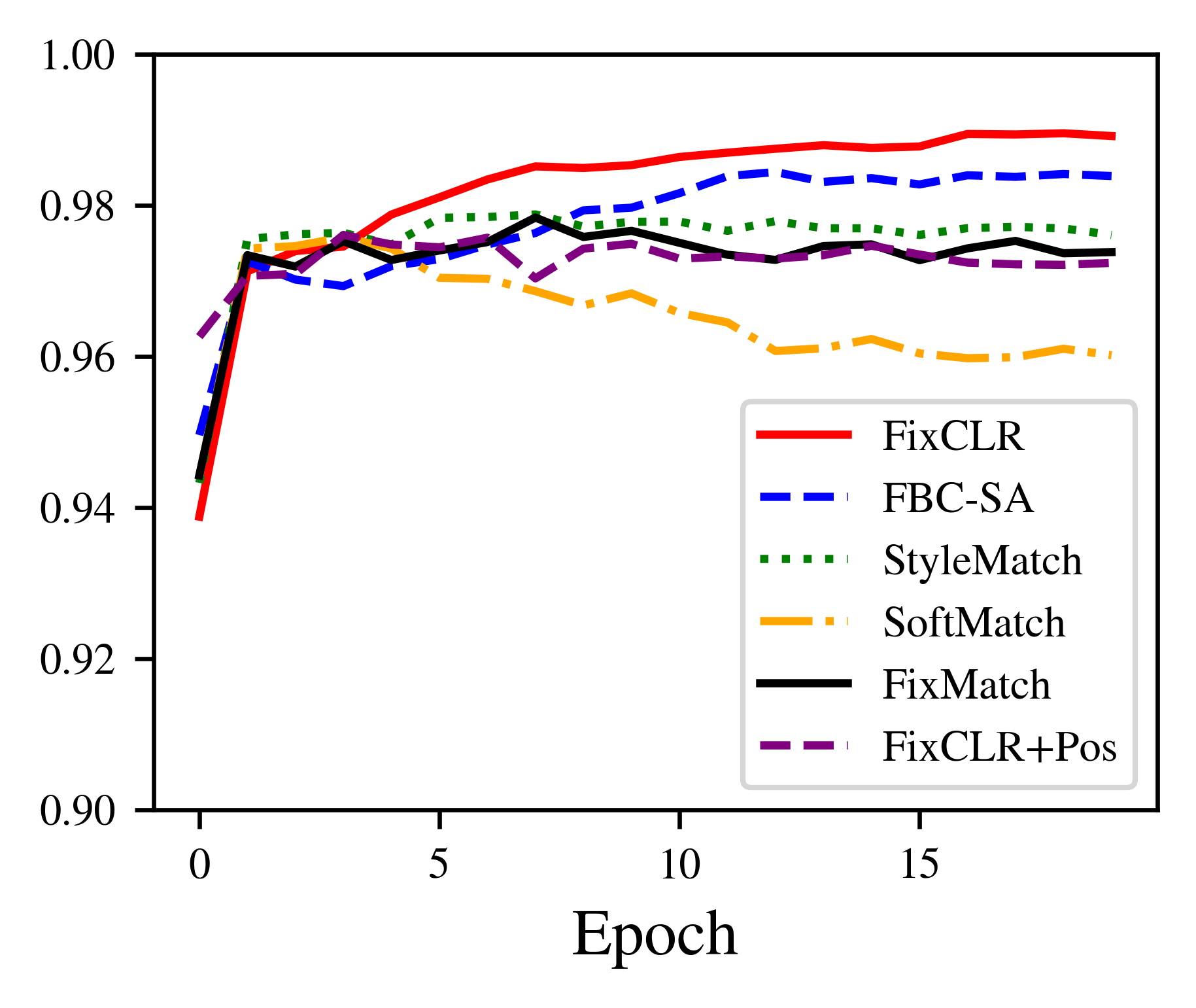}
  \caption{Pseudo-label accuracy}
  \label{fig2:sub1}
\end{subfigure}%
\begin{subfigure}{.22\textwidth}
  \centering
  \includegraphics[width=.98\linewidth,trim=3 3 3 3,clip]{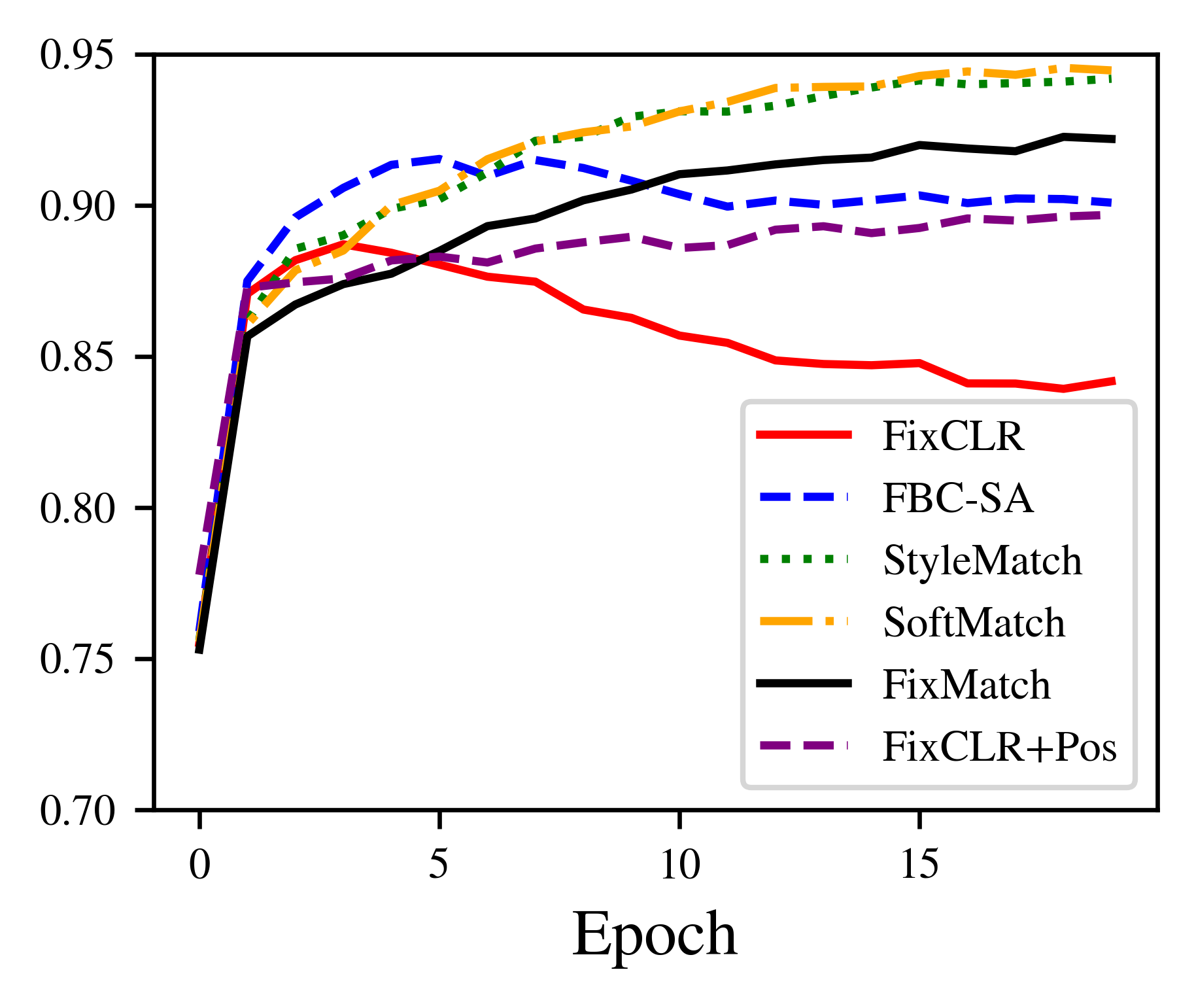}
  \caption{Pseudo-label keep ratio}
  \label{fig2:sub2}
\end{subfigure}
\caption{Pseudo-labels accuracy (quality) and keep ratio (quantity) after thresholding on the PACS dataset using an ImageNet pretrained network}
\label{fig2:quality_quantity}
\end{figure}

\noindent\textbf{Effects of FixCLR. }
FixCLR effectively regularizes the model to learn domain invariant features by minimizing the similarity between same-class groups and different-class domain groups. As shown in Figure \ref{fig1:sub4}, FixCLR avoids clustering domains, unlike other methods. This suggests that by explicitly encouraging domain invariance, FixCLR enables better performance on unseen domains, as the model can focus on features that contribute to class information regardless of domain information. Our results across various datasets support this qualitative finding. (The learned representation space of other methods are shown in the supplementary materials.)

Although FixCLR does not include a term specifically designed to improve pseudo-labelling, we study its effects on pseudo-label quality and quantity, which are important for SSDG performance. Quality is especially important for FixCLR as it depends on pseudo-labels for domain invariant regularization. Figure \ref{fig2:sub1} shows the quality of pseudo-labels, measured by accuracy after thresholding, across epochs. Accuracy is calculated based on the ground truth labels, which the model does not have access to during training. Figure \ref{fig2:sub2} shows the quantity of pseudo-labels, measured by the ratio of samples that are above the threshold, across epochs. We find that FixCLR improves pseudo-label quality while reducing quantity. 

The quality-quantity trade off of FixCLR is intuitive as FixMatch and FixCLR are seen to have different learned representational structure in Figure \ref{fig_manifold}. While FixMatch slightly clusters domains (Figure \ref{fig1:sub3}), FixCLR shows a different structure with minimal domain clusters (Figure \ref{fig1:sub4}). The structure difference suggests that the regularization of FixCLR changes the effect of cross-entropy, likely weakening its signal, which can lower model confidence and reduce quantity. Consequently, the model uses only the more confidently classifiable samples, thereby increasing quality. A similar, but less pronounced, trend is seen with FBC-SA.

\begin{table}
\centering
\fontsize{9pt}{9pt}\selectfont
\setlength{\tabcolsep}{2.0pt} 
\renewcommand{\arraystretch}{0.9} 
\begin{tabular}{l|ccc|ccc}
\toprule
\begin{tabular}[c]{@{}l@{}}Dataset/\\Method\end{tabular}  & FixCLR  
& +Soft
& +Style
& \begin{tabular}[c]{@{}l@{}}FixCLR\\w. Positive\end{tabular}
& +Soft
& +Style
\\ \midrule

PACS    & 76.0\%  & 76.8\%  & 79.0\%  & 75.5\% & 75.8\% & 78.3\% \\
Terra   & 33.8\%  & 35.5\%  & 33.5\%  & 30.8\% & 24.0\% & 29.7\% \\
IMG-R   & 24.0\%  & 25.9\%  & 24.7\%  & 21.8\% & 23.8\% & 22.1\%\\
FMOW    & 24.5\%  & 22.3\%  & 25.4\%  & 24.1\% & 19.3\% & 23.8\%  \\ 
\bottomrule
\end{tabular}
\caption{Ablation study with same-class attraction on pretrained network}
\label{tab4:fixclr_positive}
\end{table}


\noindent \textbf{Using positive attraction. }
We deliberately chose not to use positive class attraction, where the similarity is maximized between the representation of samples with the same predicted pseudo-label. 
Figure \ref{fig2:sub1} shows that positive class attraction (FixCLR+Pos.) starts with high quality pseudo-labels, as seen with the initial high value at epoch 0. This is likely because positive attraction reinforces the cross-entropy loss signal, helping the model to quickly become confident only for easier samples, but has difficulty aligning samples across domains due to their vastly different distributions. As training progresses and quantity increases (Figure \ref{fig2:sub2}), quality does not increase like FixCLR but remains slightly worse than FixMatch. 
Figure \ref{fig1:positive} shows that when positive attraction is used, domain clusters are present, unlike FixCLR (Figure \ref{fig1:sub4}).  
Accuracy is also consistently worse as shown in Table \ref{tab4:fixclr_positive}. These results support that FixCLR benefits from utilizing only the repelling term.
\textcolor{black}{Our findings align with results from a related study on fully supervised domain generalization \cite{yao2022pcl}, which reported that attracting positive pairs from vastly different distributions, as often seen in domain generalization data, harms generalization. Although their proxy-based contrastive loss requires labels and cannot be directly applied when labels are unavailable, our results with FixCLR similarly demonstrate that positive contrastive attraction of different domains is undesirable for SSDG.} 

\noindent\textbf{FixCLR as a form of Negative Learning.}
We have primarily interpreted FixCLR as a SimCLR-inspired regularization term that includes a special negative sample selection process, focusing only on the negative repelling term. FixCLR can also be viewed as a form of negative learning \cite{kim2019nlnl} that uses complementary labels. Rather than using potentially incorrect labels (e.g., `A'), it trains with complementary labels (e.g., `not-B'), which reduces risk for training with noisy labels. FixCLR groups samples by pseudo-labels and repels those from other classes, making this repelling process similar to using complementary labels. However, unlike traditional negative learning, which selects one complementary label, FixCLR repels all classes except the predicted pseudo-label. Additionally, the FixCLR loss, defined in Eq.\ref{eq:fixclr}, is a regularization term and is not the only loss function used for training. The combination of FixCLR regularization with the cross-entropy loss helps prevent underfitting, an issue observed in the original negative learning framework \cite{kim2021joint}. Consequently, FixCLR further separates confident pseudo-labels, improving pseudo-label quality and decreasing quantity, as shown in Figure \ref{fig2:quality_quantity}. The separation of confident pseudo-labels also parallels improvements to negative learning \cite{kim2021joint}, where the combination of positive and negative learning helps prevents underfitting.

\section{Conclusion}
In conclusion, we introduced FixCLR, a plug-and-play regularization method that explicitly regularizes for domain invariance for semi-supervised domain generalization (SSDG). FixCLR adapts contrastive learning to be used for SSDG by changing two crucial components: using class information from pseudo-labels and using only a repelling term. 
Our extensive experiments demonstrate that FixCLR is a simple yet highly effective regularization for improving accuracy in SSDG settings, especially in combination with other methods. 
Overall, our findings emphasize the importance of explicitly regularizing for domain invariance and considering both pseudo-label quality and quantity. Lastly, we caution the usage of pretrained models in SSDG due to potential domain information leakage.

{
    \small
    \bibliographystyle{ieeenat_fullname}
    \bibliography{main}
}

\clearpage
\setcounter{page}{1}
\appendix

\maketitlesupplementary

\section{5-Label Setting}
In the main text, we discussed experiments using a 10-label setting. Following the work of \citet{galappaththige2024towardsfbc} and \citet{zhou2023semistyle}, we also conducted experiments with a 5-label setting. This setting is more challenging due to the reduced number of available labels, which increases the importance of semi-supervised methods.

As shown in Tables \ref{tab:5Lab_highest_pre} and \ref{tab:5Lab_highest_non}, we observed that a combination of FixCLR and StyleMatch consistently performed the best. This is particularly evident in the non-pretrained case, where FixCLR combined with StyleMatch achieved the highest performance in 5 out of 6 datasets. This improvement is likely due to StyleMatch's use of a pretrained network for style augmentation. This may involve some domain information leakage, as the model might have been exposed to images similar to the target domain during pretraining, as discussed in the main text.

Nevertheless, our conclusions remain unchanged: FixCLR is an effective plug-and-play regularization technique for ensuring domain invariance across all domains, especially when used with other methods. For more detailed results on the 5-label setting, refer to Tables \ref{tab:5Lab_pretrained}a and \ref{tab:5Lab_nonpretrained}.

\begin{table}[h!]
\centering
\fontsize{9pt}{9pt}\selectfont
\setlength{\tabcolsep}{1.15pt} 
\renewcommand{\arraystretch}{1.0} 
\begin{tabular}{l|cccccc}
\toprule
Dataset & Digits  & PACS  & OH  & Terra  & IMG-R & FMOW \\
 \midrule
\begin{tabular}[l]{@{}l@{}}Best \\Method\end{tabular}  & 
\begin{tabular}[c]{@{}l@{}}FixCLR \\+ Soft\end{tabular} & 
\begin{tabular}[c]{@{}l@{}}FixCLR \\+ Style\end{tabular} & 
\begin{tabular}[c]{@{}l@{}}FixCLR \\+ Soft\end{tabular} & 
\begin{tabular}[c]{@{}l@{}}FixCLR \\+ Soft\end{tabular} & 
\begin{tabular}[c]{@{}l@{}}FixCLR \\+ Style\end{tabular} & 
\begin{tabular}[c]{@{}l@{}}FixCLR \\+ Style\end{tabular}  \\

\\

Accuracy & 61.3 & 77.9 & 57.1 & 25.0 & 25.9 & 25.3 
\\ \bottomrule
\end{tabular}
\caption{Highest performing method on the ImageNet pretrained network (5 labels per class)}
\label{tab:5Lab_highest_pre}
\end{table}

\begin{table}[h!]
\centering
\fontsize{9pt}{9pt}\selectfont
\setlength{\tabcolsep}{1.15pt} 
\renewcommand{\arraystretch}{1.0} 
\begin{tabular}{l|cccccc}
\toprule
Dataset & Digits  & PACS  & OH  & Terra  & IMG-R & FMOW \\
 \midrule
\begin{tabular}[l]{@{}l@{}}Best \\Method\end{tabular}  & 
\begin{tabular}[c]{@{}l@{}}FBC-SA \\+ Soft\end{tabular} & 
\begin{tabular}[c]{@{}l@{}}FixCLR \\+ Style\end{tabular} & 
\begin{tabular}[c]{@{}l@{}}FixCLR \\+ Style\end{tabular} & 
\begin{tabular}[c]{@{}l@{}}FixCLR \\+ Style\end{tabular} & 
\begin{tabular}[c]{@{}l@{}}FixCLR \\+ Style\end{tabular} & 
\begin{tabular}[c]{@{}l@{}}FixCLR \\+ Style\end{tabular}  \\

\\

Accuracy & 54.8  & 31.3  & 18.7 & 22.0 & 8.3 &  16.8
\\ \bottomrule
\end{tabular}
\caption{Highest performing method on the non-pretrained network (5 labels per class)}
\label{tab:5Lab_highest_non}
\end{table}

\newpage

\section{Learned Representation Space}
\begin{figure*}[h!]
\centering
\begin{subfigure}{.19\textwidth}
  \centering
  \includegraphics[width=.95\linewidth,trim=2 2 2 2,clip]{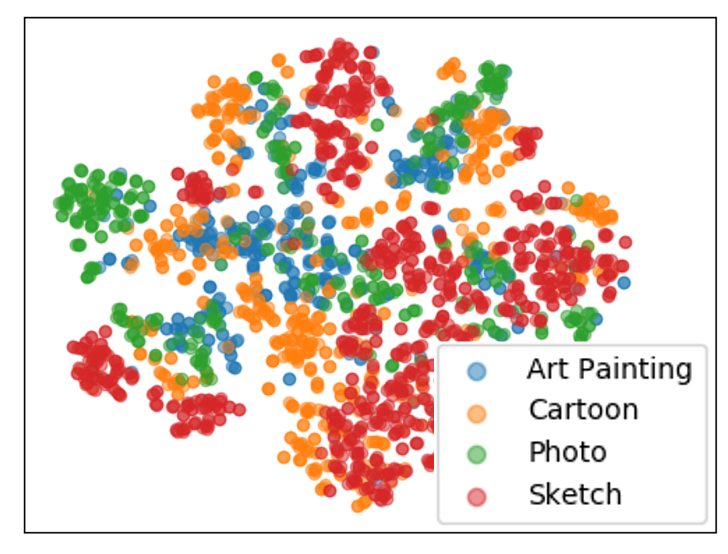}
  \caption{FixMatch}
  \label{a_fig:sub1}
\end{subfigure}%
\begin{subfigure}{.19\textwidth}
  \centering
  \includegraphics[width=.95\linewidth,trim=2 2 2 2,clip]{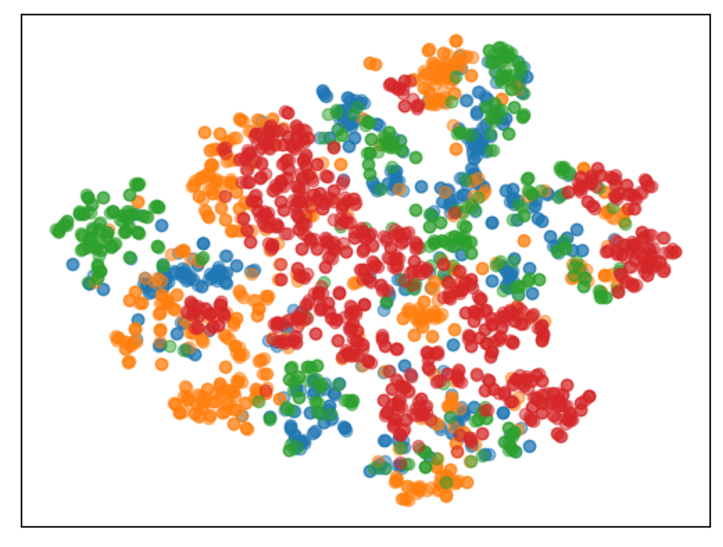}
  \caption{SoftMatch}
  \label{a_fig:sub3}
\end{subfigure}
\begin{subfigure}{.19\textwidth}
  \centering
  \includegraphics[width=.95\linewidth,trim=2 2 2 2,clip]{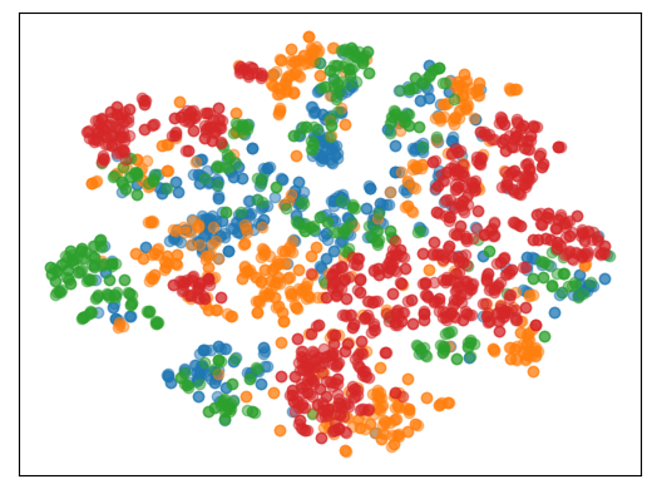}
  \caption{StyleMatch}
  \label{a_fig:sub3}
\end{subfigure}
\begin{subfigure}{.19\textwidth}
  \centering
  \includegraphics[width=.95\linewidth,trim=2 2 2 2,clip]{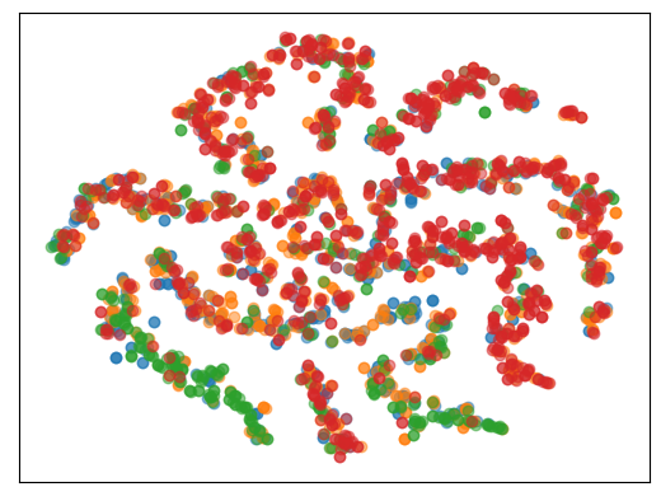}
  \caption{FBC-SA}
  \label{a_fig:sub3}
\end{subfigure}
\begin{subfigure}{.19\textwidth}
  \centering
  \includegraphics[width=.95\linewidth,trim=2 2 2 2,clip]{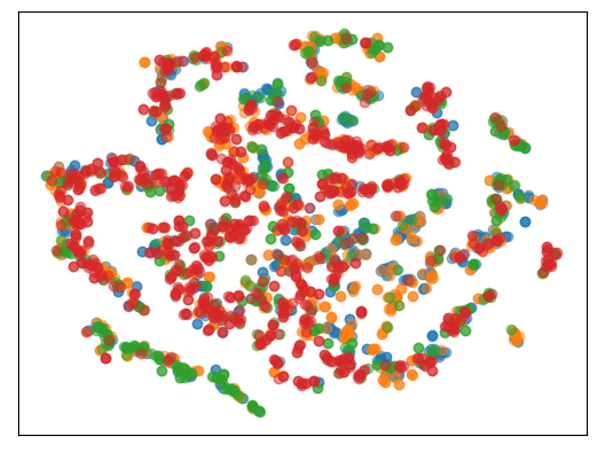}
  \caption{FixCLR}
  \label{a_fig:sub3}
\end{subfigure}

\begin{subfigure}{.19\textwidth}
  \centering
  \includegraphics[width=.95\linewidth,trim=2 2 2 2,clip]{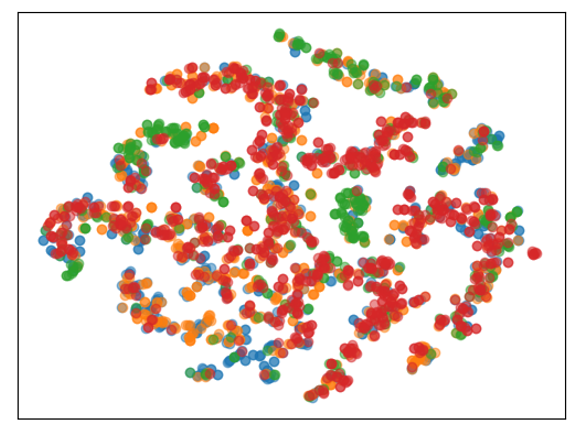}
  \caption{Soft+FBC-SA}
  \label{a_fig:sub3}
\end{subfigure}
\begin{subfigure}{.19\textwidth}
  \centering
  \includegraphics[width=.95\linewidth,trim=2 2 2 2,clip]{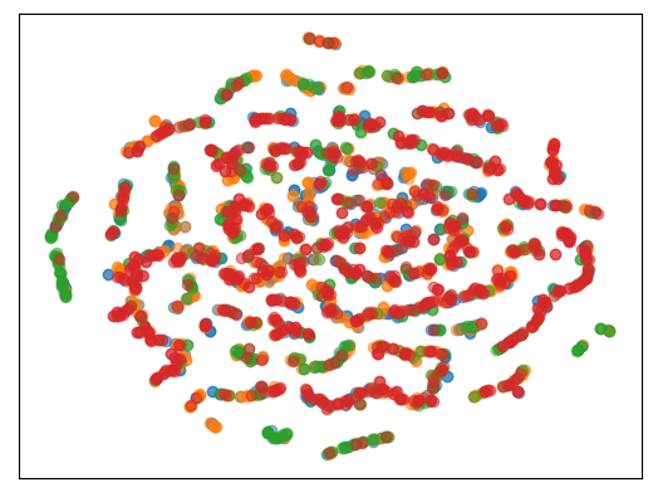}
  \caption{Soft+FixCLR}
  \label{a_fig:sub3}
\end{subfigure}
\begin{subfigure}{.19\textwidth}
  \centering
  \includegraphics[width=.95\linewidth,trim=2 2 2 2,clip]{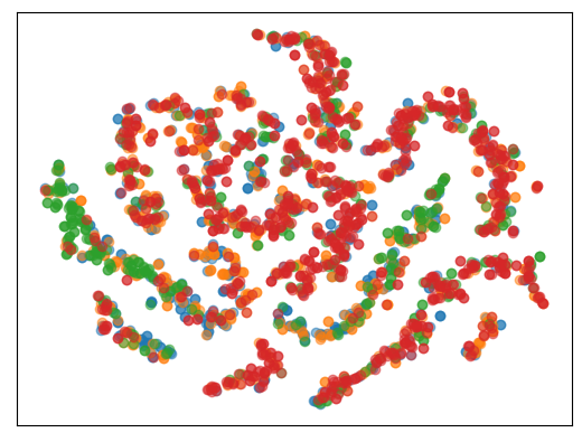}
  \caption{Style+FBC-SA}
  \label{a_fig:sub3}
\end{subfigure}
\begin{subfigure}{.19\textwidth}
  \centering
  \includegraphics[width=.95\linewidth,trim=2 2 2 2,clip]{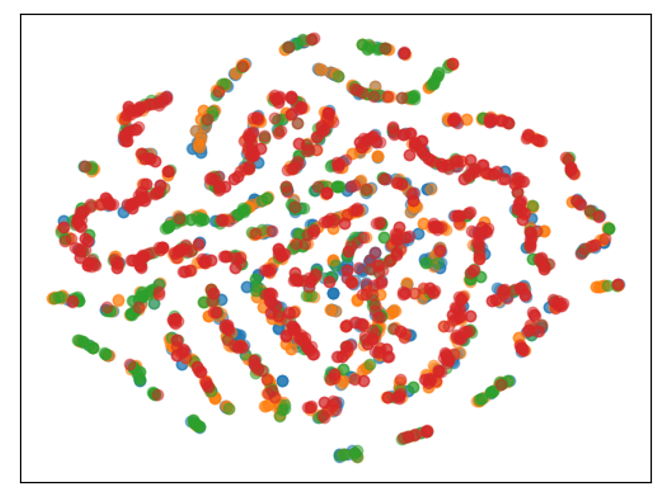}
  \caption{Style+FixCLR}
  \label{a_fig:sub3}
\end{subfigure}

\caption{t-SNE of the learned representation space of the learned encoder (\textbf{ImageNet pretrained}) on the PACS dataset (10 labels per class) when trained with all domains. The colors represent different domains.}
\label{fig_all_manifold}
\end{figure*}

We also present the learned representation spaces for various methods and their combinations. Figure \ref{fig_all_manifold} displays the representation space for the pretrained model. The results are intuitive: FixMatch, SoftMatch, and StyleMatch do not explicitly enforce domain invariance, leading to visible domain clusters in the learned representation space. The cross-entropy loss does not encourage domain-invariant representations since it can still accurately classify samples, as mentioned in the main text. In contrast, FBC-SA, FixCLR, and their combinations show no visible domain clustering.

\begin{figure*}[h!]
\centering
\begin{subfigure}{.19\textwidth}
  \centering
  \includegraphics[width=.95\linewidth,trim=2 2 2 2,clip]{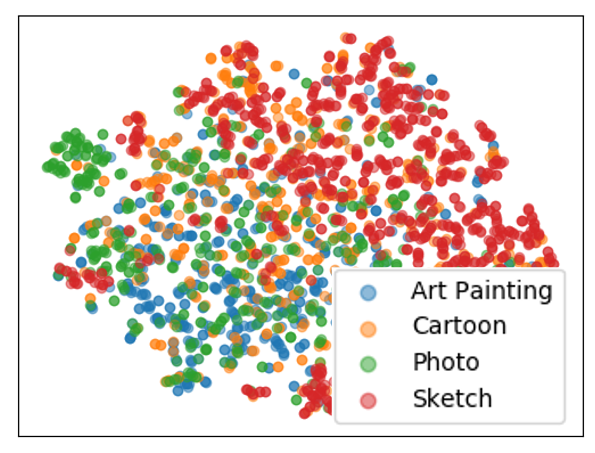}
  \caption{FixMatch}
  \label{a_fig:sub1}
\end{subfigure}%
\begin{subfigure}{.19\textwidth}
  \centering
  \includegraphics[width=.95\linewidth,trim=2 2 2 2,clip]{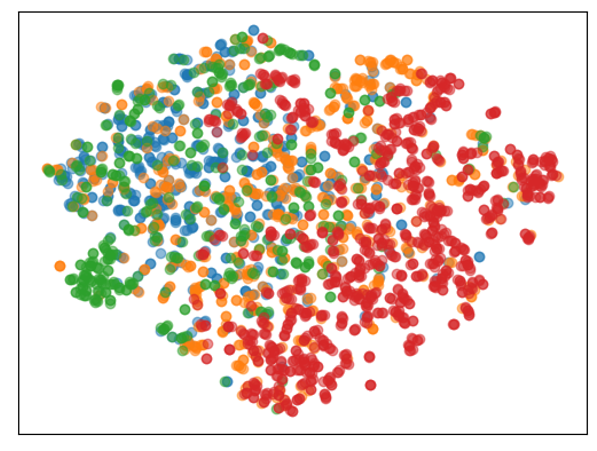}
  \caption{SoftMatch}
  \label{a_fig:sub3}
\end{subfigure}
\begin{subfigure}{.19\textwidth}
  \centering
  \includegraphics[width=.95\linewidth,trim=2 2 2 2,clip]{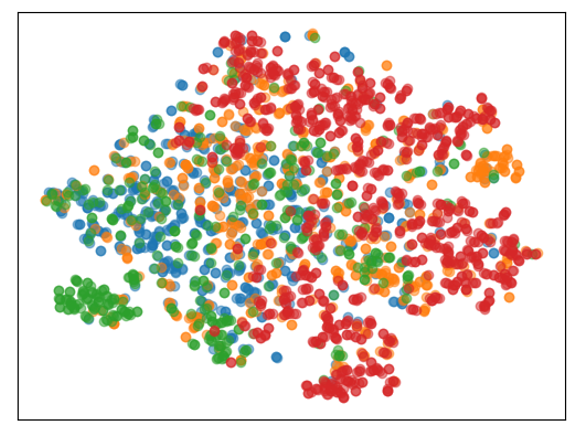}
  \caption{StyleMatch}
  \label{a_fig:sub3}
\end{subfigure}
\begin{subfigure}{.19\textwidth}
  \centering
  \includegraphics[width=.95\linewidth,trim=2 2 2 2,clip]{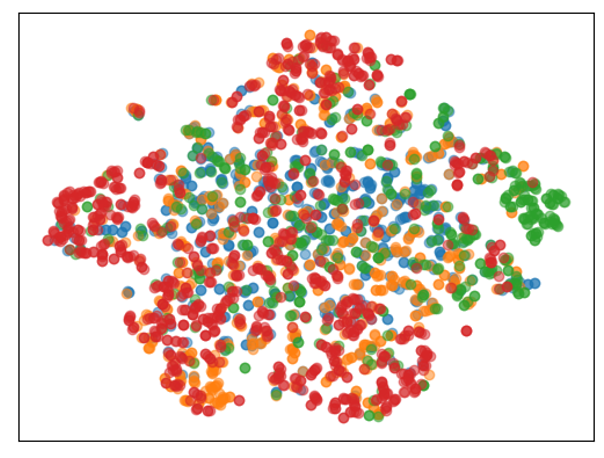}
  \caption{FBC-SA}
  \label{a_fig:sub3}
\end{subfigure}
\begin{subfigure}{.19\textwidth}
  \centering
  \includegraphics[width=.95\linewidth,trim=2 2 2 2,clip]{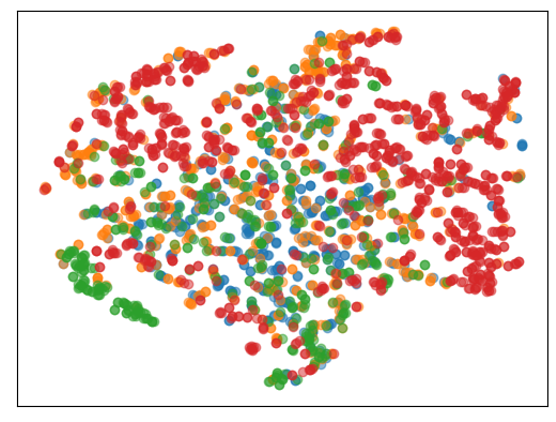}
  \caption{FixCLR}
  \label{a_fig:sub3}
\end{subfigure}

\begin{subfigure}{.19\textwidth}
  \centering
  \includegraphics[width=.95\linewidth,trim=2 2 2 2,clip]{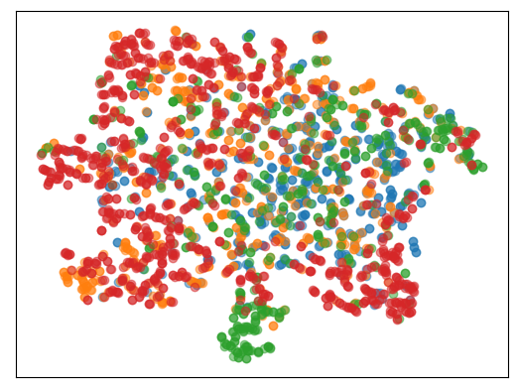}
  \caption{Soft+FBC-SA}
  \label{a_fig:sub3}
\end{subfigure}
\begin{subfigure}{.19\textwidth}
  \centering
  \includegraphics[width=.95\linewidth,trim=2 2 2 2,clip]{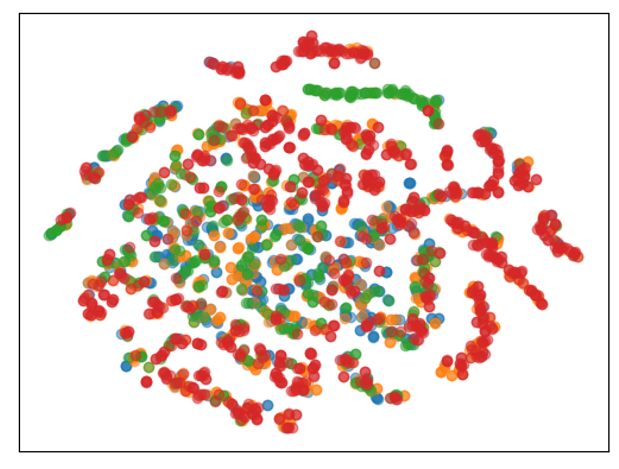}
  \caption{Soft+FixCLR}
  \label{a_fig:sub3}
\end{subfigure}
\begin{subfigure}{.19\textwidth}
  \centering
  \includegraphics[width=.95\linewidth,trim=2 2 2 2,clip]{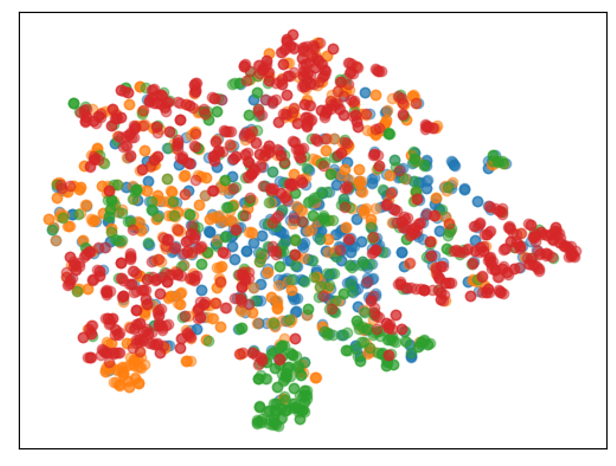}
  \caption{Style+FBC-SA}
  \label{a_fig:sub3}
\end{subfigure}
\begin{subfigure}{.19\textwidth}
  \centering
  \includegraphics[width=.95\linewidth,trim=2 2 2 2,clip]{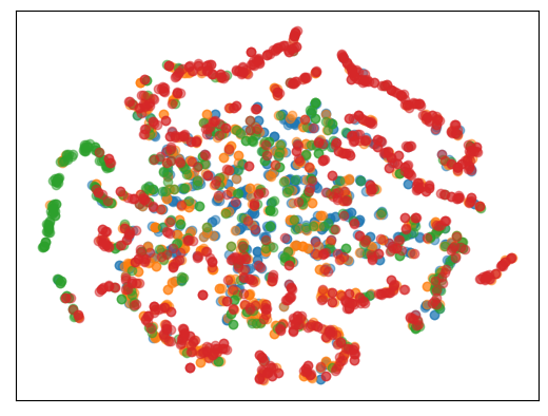}
  \caption{Style+FixCLR}
  \label{a_fig:sub3}
\end{subfigure}

\caption{t-SNE of the learned representation space of the learned encoder \textbf{(non-pretrained)} on the PACS dataset (10 labels per class) when trained with all domains. The colors represent different domains.}
\label{fig_all_manifold_non}
\end{figure*}

On the PACS dataset, FBC-SA and FixCLR have similar performance, though FixCLR generally excels on more complex datasets with additional domains, which are harder to visualize clearly. An interesting observation is presented in Figure \ref{fig_all_manifold_non} for the non-pretrained network. None of the standalone methods achieve domain-invariant representations, as indicated by the presence of domain clusters. However, FixCLR, when combined with SoftMatch and StyleMatch, shows no domain clusters, unlike combinations with FBC-SA. This further supports that FixCLR is particularly effective at enforcing domain invariance across all domains.

\begin{table*}
\renewcommand{\arraystretch}{1.15}

\centering
\fontsize{9pt}{9pt}\selectfont
\begin{tabular}{cccccccc} 
\toprule
~                           & Method / Dataset     & Digits DG & PACS & Office-Home   & Terra Inc. & ImageNet-R & FMOW-Wilds  \\ 
\hline
~ & FixMatch   & 48.8   & 72.0 & 53.5 & 20.5 & 24.0 & \underline{24.3}  \\ 
\hline
\multirow{2}{*}{Quality}    & DebiasPL   & 47.5   & 69.8 & 53.0 & 20.3 & 23.8 & 21.0  \\
                            & DeFixMatch & 21.3   & 76.0 & 55.3 & 14.5 & \underline{24.8} & 21.5  \\ 
\hline
\multirow{2}{*}{Quantity}   & FlexMatch  & 50.3   & 75.8 & 56.3   & \underline{22.3} & 23.3 & 21.4  \\
                            & FreeMatch  & \underline{58.5}   & 76.5 & 54.5   & \underline{22.3} & 23.8 & 24.2  \\ 
\hline
Quality + Quantity          & SoftMatch  & 52.1   & 74.0 & 55.0   & 19.3 & 24.1 & 19.2  \\ 
\hline
\multirow{3}{*}{SSDG}       & StyleMatch & 48.0   & \underline{77.5} & \underline{56.5}   & 20.0 & 24.5 & 22.0  \\
                            & FBC-SA     & 48.8   & 73.3 & 55.3   & 16.5 & 23.3 & -     \\
                            & FixCLR     & 52.3   & 73.8 & 54.3   & 20.5 & 24.0 & 19.2  \\ 
\hline \hline
\multicolumn{8}{c}{Improvements when FBC-SA/FixCLR added}                     \\ 
\hline
\multirow{2}{*}{DebiasPL}   & +FBC-SA    & +8.0  & +\textbf{4.0 }& +3.3  & +2.2  & -8.0 & -     \\
                            & +FixCLR    & \textbf{+12.0} & +2.3 & \textbf{+3.5}  & \textbf{+7.0}  & \textbf{+1.5} & 0.2     \\
\hline
\multirow{2}{*}{DeFixMatch} & +FBC-SA    & \textbf{-3.8} & -1.7 & \textbf{+1.2}  & +1.3  & \textbf{-0.9} & -     \\
                            & +FixCLR    & -4.5 & \textbf{-1.2} & +1.0  & \textbf{+7.8}  & -1.6 & -3.2     \\
\hline
\multirow{2}{*}{FlexMatch}  & +FBC-SA    & \textbf{+1.5} & -1.8 & -1.0  & -2.5  & -5.9 & -6.7     \\
                            & +FixCLR    & +0.7 & \textbf{-1.5} & \textbf{-0.5}  & \textbf{+2.5}  & \textbf{+1.6} & \textbf{-2.7}     \\
\hline
\multirow{2}{*}{FreeMatch}  & +FBC-SA    & -7.0 & -3.0 & \textbf{+0.8} & -2.5 & +0.0 & -9.2     \\
                            & +FixCLR    & \textbf{-4.7} & \textbf{-2.2} & +0.4 & \textbf{+1.2} & \textbf{+0.5} & \textbf{-3.6}     \\
\hline
\multirow{2}{*}{SoftMatch}  & +FBC-SA    & +8.4 & \textbf{+2.3} & +1.5 & -2.0 & -0.7 & -     \\
                            & +FixCLR    & \textbf{+9.2} & +0.9 & \textbf{+2.1} & \textbf{+5.7} & \textbf{+1.8} & -1.2     \\
\midrule
\multirow{2}{*}{StyleMatch} & +FBC-SA    & +4.0 & +0.3 & +0.0 & +0.5 & +0.1 & -   \\
                            & +FixCLR    & \textbf{+4.5} & \textbf{+0.4} & \textbf{+0.5} & \textbf{+0.8} & \textbf{+0.2} & +3.3  \\
\bottomrule
\end{tabular}
\caption{Performance on various datasets (5 labels per class) with a \textbf{non-pretrained} network. We \underline{underline} the best \underline{\textbf{standalone}} method, and \textbf{bold} the best improvement for each method when FBC-SA / FixCLR are added.}
\label{tab:5Lab_pretrained}
\end{table*}

\begin{table*}
\renewcommand{\arraystretch}{1.15}

\centering
\fontsize{9pt}{9pt}\selectfont
\begin{tabular}{cccccccc} 
\toprule
~                           & Method / Dataset     & Digits DG & PACS & Office-Home   & Terra Inc. & ImageNet-R & FMOW-Wilds  \\ 
\hline
~ & FixMatch   & 42.0   & 28.8   & 16.8 & 17.3 & 6.9 & 15.1  \\ 
\hline
\multirow{2}{*}{Quality}    & DebiasPL   & 44.3   & 29.3 & 15.8 & 16.8 & 7.4 & 14.8  \\
                            & DeFixMatch & 19.8   & 28.8 & 16.0 & 16.0 & 6.7 & 12.3  \\ 
\hline
\multirow{2}{*}{Quantity}   & FlexMatch  & 46.5   & 28.8 & 16.3   & 16.5 & 7.1 & 15.3  \\
                            & FreeMatch  & \underline{48.8}   & 30.0 & 16.0   & 17.0 & 7.0 & 15.3  \\ 
\hline
Quality + Quantity          & SoftMatch  & 48.3   & \underline{32.3} & 17.3   & 15.8 & 6.8 & 16.2  \\ 
\hline
\multirow{3}{*}{SSDG}       & StyleMatch & 44.8   & 29.8 & \underline{17.5}   & \underline{19.5} & \underline{7.9} & \underline{16.6}  \\
                            & FBC-SA     & 47.8   & 30.8 & 15.8   & 15.8 & 7.4 & 15.5     \\
                            & FixCLR     & 45.0   & 28.5 & 17.0   & 19.3 & 7.2 & 15.9  \\ 
\hline \hline
\multicolumn{8}{c}{Improvements when FBC-SA/FixCLR added}                     \\ 
\hline
\multirow{2}{*}{DebiasPL}   & +FBC-SA    & \textbf{+3.0} & +0.0 & +0.2  & -3.5 & +0.1 & -0.1     \\
                            & +FixCLR    & +0.5 & +0.0 & \textbf{+0.5}  & \textbf{-1.8} & -0.4 & \textbf{+1.4}     \\
\hline
\multirow{2}{*}{DeFixMatch} & +FBC-SA    & -4.3 & +0.0 & +0.0 & \textbf{+1.5} & +0.5 & -0.9     \\
                            & +FixCLR    & \textbf{+0.7} & \textbf{+0.7} & +0.0 & +1.0 & +0.6  & \textbf{+4.1}     \\
\hline
\multirow{2}{*}{FlexMatch}  & +FBC-SA    & -1.7 & +0.7 & -0.5 & -1.0 & -0.1 & +0.4     \\
                            & +FixCLR    & \textbf{+2.5} & \textbf{+2.0} & -0.3 & \textbf{+1.3} & \textbf{+0.3} & \textbf{+0.9}     \\
\hline
\multirow{2}{*}{FreeMatch}  & +FixCLR    & \textbf{-1.5} & \textbf{-0.5} & \textbf{+0.8} & \textbf{+1.8} & +0.5 & \textbf{+1.2}   \\
                            & +FBC-SA    & -4.0 & -1.5 & +0.5 & -1.5 & \textbf{+0.6} & -0.4     \\
\hline
\multirow{2}{*}{SoftMatch}  & +FBC-SA    & \textbf{+6.5} & -3.5 & +0.5  & -2.8 & +0.2 & -0.8     \\
                            & +FixCLR    & +3.7 & \textbf{-2.0} & +0.5  & \textbf{+0.5} & \textbf{+0.5} & \textbf{+0.3}     \\\midrule
\multirow{2}{*}{StyleMatch} & +FBC-SA    & -7.0 & -2.0 & +1.0  & -2.0 & +0.0 & -2.0     \\
                            & +FixCLR    & \textbf{+2.5} & \textbf{+1.5} & \textbf{+1.2 } & \textbf{+2.5} & \textbf{+0.4 }& \textbf{+0.2}     \\
\bottomrule
\end{tabular}
\caption{Performance on various datasets (5 labels per class) with a \textbf{non-pretrained} network. We \underline{underline} the best \underline{\textbf{standalone}} method, and \textbf{bold} the best improvement for each method when FBC-SA / FixCLR are added.}
\label{tab:5Lab_nonpretrained}
\end{table*}

\end{document}